\documentclass[11pt,a4paper]{article}
\usepackage[utf8]{inputenc}
\usepackage{hyperref}
\usepackage{todonotes}
\usepackage[nohyperref]{udst2018}
\usepackage{times}
\usepackage{latexsym}
\usepackage{natbib}

\usepackage{multirow}
\usepackage{url}
\usepackage{microtype}
\usepackage{paralist}
\usepackage{booktabs}
\usepackage{covington}
\usepackage{verbatim}
\usepackage{amsmath}
\usepackage{csquotes}
\usepackage{algorithm}
\usepackage{algorithmic}
\usepackage[skip=4pt]{caption}

\aclfinalcopy 
\def\aclpaperid{8} 

\newcommand{\lang}[1]{\textit{#1}}
\newcommand{\action}[1]{\texttt{#1}}

\title{Finding the way from ä to a:\\ Sub-character morphological inflection for the SIGMORPHON 2018 Shared Task}

\author{Fynn Schröder\thanks{\enspace These authors contributed equally} \qquad Marcel Kamlot\footnotemark[1] \qquad Gregor Billing\footnotemark[1] \qquad Arne Köhn \\
  Department of Computer Science \\ Universität Hamburg \\ Germany \\
  {\tt \{%
\href{mailto:7schroed@informatik.uni-hamburg.de}{7schroed},%
\href{mailto:6kamlot@informatik.uni-hamburg.de}{6kamlot},%
\href{mailto:5billing@informatik.uni-hamburg.de}{5billing},%
\href{mailto:koehn@informatik.uni-hamburg.de}{koehn}%
\}@informatik.uni-hamburg.de}\\}

\date{}

\begin{document}
\maketitle
\newcommand{\prostst}[0]{\emph{CoNLL--SIGMORPHON 2018 Shared Task}}

\begin{abstract}
In this paper we describe the system submitted by UHH to the \prostst: Universal Morphological Reinflection. We propose a neural architecture based on the concepts of UZH \cite{cluzh:MakarovRC17}, adding new ideas and techniques to their key concept and evaluating different combinations of parameters. The resulting system is a language-agnostic network model that aims to reduce the number of learned edit operations by introducing equivalence classes over graphical features of individual characters. We try to pinpoint advantages and drawbacks of this approach by comparing different network configurations and evaluating our results over a wide range of languages.
\end{abstract}

\section{Introduction}

The system described in this paper\footnote{Source code available at \url{https://gitlab.com/nats/sigmorphon18}} was submitted for the \prostst~\citep{sigmorphon:st2018}, part 1 only. This assignment challenges the participants to design systems that generate inflected forms based on an input lemma and feature set as shown in \autoref{fig:example-inflection}.

Training data is usually provided in three different volumes (see \autoref{fig:training-volumes}), all conforming to the \textit{UniMorph} standard proposed by \citet{kirov:unimorph}. The entire data set comprises 103 languages, although not every training volume is available for every language. In addition, some languages have significantly less training samples than the maximum depicted in \autoref{fig:training-volumes}. 

\begin{figure}
\centering
\begin{tabular}{cc}
bungas & \texttt{N;INST;PL}\\  
\multicolumn{2}{c}{$\Downarrow$}\\ \addlinespace
\multicolumn{2}{c}{bungām}
\end{tabular}
\caption{An example for word inflection in \lang{Latvian}, "a drum/drums"}
\label{fig:example-inflection}
\end{figure}

\begin{table}
\centering
\begin{tabular}{lrr}
\toprule
& \multicolumn{2}{c}{\# of Samples}\\ \cmidrule{2-3}
Volume & max & avg\\
\midrule
low & 100 & 99.6\\
medium & 1.000 & 934.5\\
high & 10.000 & 8553.6\\
\bottomrule
\end{tabular}
\caption{Maximum training data volumes}
\label{fig:training-volumes}
\end{table}

With such a high count of diverse languages, our system is not tailored towards specific linguistic features of a language, but instead learns transition-based character actions to transform a lemma into its inflected form. We try to limit the number of output actions that our network has to learn by grouping certain characters into common groups based on graphical features like accents or symbol modifiers. 
Lastly, we propose a method to enhance the training data of the low setting without the use of external resources.

\section{String Transducer}
\label{sec:transducer}
The inflection process itself is realized in our system through a finite set of edit actions, resulting in a standard transducer process. An input string is traversed left-to-right via an index pointer that indicates which symbol is currently being regarded.
The following actions are available:
\begin{compactitem}
	\item \action{EMIT} $s$ (for any symbol $s$): Appends $s$ to the output string, irrespective of pointer symbol
    \item \action{COPY}: Append the pointer symbol to the output string
    \item \action{PATCH} $x$: Apply the graphical patch matrix $x$ (cf. Section~\ref{sec:patches}) to the pointer symbol and append the result to the output string
    \item \action{MOVE}: Increment the pointer to continue traversing the input word
    \item \action{EOW} (\underline{e}nd \underline{o}f \underline{w}ord): Stop traversing the string and consider the current output string as the final inflection result
\end{compactitem}

\subsection{Alignment}
We chose to implement our own mechanism to align input lemma and output strings, to accommodate for our patch concept.

The aligner itself is based on plain Levenshtein metrics \citep{levenshtein:binary66}, with the additional constraint that two symbols $a, b$ are considered equal (cost $0$) if there is a patch that transforms $a$ into $b$. We then pick the alignment with the lowest cost according to this customized Levenshtein metric to encourage our system to learn \action{COPY} and \action{PATCH} actions as much as possible.

\subsection{Oracle Algorithm}
\label{sec:oracle-alg}
The actions needed to transform an input lemma $w$ into the inflected target form $t$ are generated through a deterministic algorithm that acts as static oracle gold standard. This algorithm works with an aligned pair $(w', t')$ as input, where the original $w$ and $t$ are filled with arbitrary characters not appearing in the original strings. The exact procedure can be seen in algorithm \ref{alg:oracle} with \texttt{\#}-symbols being used as gap fill characters.

\begin{algorithm}
\begin{algorithmic}
\FORALL{$(c_w, c_t)$ \textbf{in} alignment}
	\IF{$c_w = \#$}
    	\STATE $actions$.append(\action{EMIT} $c_t$)
    \ELSIF{$c_t = \#$}
    	\STATE $actions$.append(\action{MOVE})
    \ELSIF{$c_w = c_t$}
    	\STATE $actions$.append(\action{COPY})
        \STATE $actions$.append(\action{MOVE})
    \ELSIF{$patchtable$.contains($c_w, c_t$)}
    	\STATE $actions$.append(\action{PATCH} $c_w$ to $c_t$)
        \STATE $actions$.append(\action{MOVE})
    \ELSIF{$c_w \neq c_t$}
    	\STATE $actions$.append(\action{EMIT} $c_t$)
        \STATE $actions$.append(\action{MOVE})
    \ENDIF
\ENDFOR
\STATE $actions$.append(\action{EOW})
\RETURN{$actions$}
\end{algorithmic}
\caption{Deriving oracle actions gold standard from aligned input strings}
\label{alg:oracle}
\end{algorithm}

\section{Patches}
\label{sec:patches}
\begin{figure}[htb]
\centering
\includegraphics[width=\linewidth]{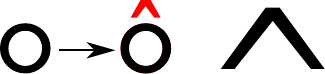}
\caption{Example patch generated from \texttt{o} to \texttt{ô} (on the right)}
\label{fig:example-patch}
\end{figure}

An essential part of our system concept is to introduce so-called \textit{patches} that act as string transducer actions. A \textit{patch} in this context is a shortcut operation between two graphically similar characters (see \autoref{fig:example-patch}), like the acute accent that transforms the letter \texttt{a} into the letter \texttt{á}. It acts as a partial function $p(x)$, so that the same patch can be applied to the letter \texttt{e} to yield $p($\texttt{e}$)=$ \texttt{é}  --- however it does not produce a valid result character when applied to the letter \texttt{b} for example. 

\subsection{Idea and Motivation}
The basic idea for these patches comes from the tendency of some languages to slightly modify the root of the word during inflection. This can either be due to phonological requirements \citep{kendris:cedilla2001} or historical linguistic influences \citep{wiese:umlaut2009,wunderlich:umlaut1999}. Two examples for inflection in \lang{German} (note the added \textit{Umlaut} symbols for the inflected forms) and \lang{French} (with added \textit{cedilla} marks) can be seen in \autoref{fig:german-declension} and \ref{fig:french-conjugation}, respectively.
The underlying intention is to capture this modification to the word stem while retaining the idea that it still is based on the same letter or group of letters. A plain transducer would identify \texttt{n} and \texttt{ñ} as different symbols, and consequently generate \action{EMIT} actions the same way it would for \texttt{f} and \texttt{g}.

\begin{table}
\centering
\begin{tabular}{llr}
\toprule
Lemma & Inflection & Features\\
\midrule
Baumhaus & Baumhäuser & N;ACC;PL\\
Kanarienvogel & Kanarienvögeln & N;DAT;PL\\
Milchkuh & Milchkühen & N;DAT;PL\\
\bottomrule
\end{tabular}
\caption{\lang{German} noun declension examples: tree house, Canary bird, (milk-)cow}
\label{fig:german-declension}

\bigbreak 

\begin{tabular}{llr}
\toprule
Lemma & Inflection & Features\\
\midrule
chacer & chaçons & V;POS;IMP;1;PL\\
évincer & évinçant & V.PTCP;PRS\\
concevoir & conçusse & V;SBJV;PST;1;SG\\
\bottomrule
\end{tabular}
\caption{\lang{French} verb conjugation examples: to hunt, to cut up, to conceive (of)}
\label{fig:french-conjugation}
\end{table}

Another motivation was the previous work performed on machine translation systems by \citet{cjk-mt:LiuLLN17}. They achieved promising results by exploring visual features on the sub-character level for machine translation, and their ideas and implementations proved useful as a starting ground for the concept presented in this section.

\subsection{Generation}
To calculate meaningful patches, we render all unique and distinct symbols contained in a given training set into binary 2D pixel matrices that contain information whether a pixel is set/black or not. The resulting matrices are then compared with an element-wise \texttt{XOR} operation that yields all pixels different between the two images. We furthermore only consider patch matrices that are based on the same ASCII character and that don't surpass a certain heuristic threshold of set pixels. Through these checks, patches from i.e. \texttt{x} to \texttt{m} get discarded because although possible, it does not produce any advantage to use them in the transducing component. The resulting effect would be the exact same as a straight-forward \action{EMIT} action.

The only non-intuitive heuristic involves the letter \texttt{i}, which contains a dot on top of a vertical bar that "disappears" when applying typical patches like accents. To counter this effect, we introduced a hard-coded set of replacement rules where the letter \texttt{i} is effectively replaced by the Turkish dotless \texttt{ı} in graphical representations, in order to fool the system into correctly applying modifications.
A similar principle might apply to other symbols in languages unknown to the authors, so the proposed architecture is capable of extending to more symbol exceptions if desired.

\subsection{NFD Unicode Decomposition}
\label{sec:nfd}
The Unicode standard proposes normalization forms\footnote{see \url{http://www.unicode.org/reports/tr15/}} that are capable of converting between composite symbols and their integral parts. In particular, the \textit{NFD} normalization achieves an effect very similar to our patch concept.

However, when designing the system we consciously decided against the use of such a feature, mostly because we were not aware of the complex NFD standard and coding a similar system by hand was not a viable alternative at all.

\subsection{Font Choice and Rendering}
\label{sec:font-choice}
The font choice for our system has to focus on two main aspects:
\begin{compactenum}
\item \label{enum:font-monospace} It has to always render all characters in the exact same position
\item \label{enum:font-unicode} It should have high Unicode coverage to be able to render as many foreign alphabets' symbols as possible
\end{compactenum}
Regarding point \ref{enum:font-monospace}, we only considered mono-space fonts and examined 14 of them.  Most of them were appropriate, only two of them still had issues with pixel-perfect alignment of the target symbols on several occasions.
Regarding Point \ref{enum:font-unicode}, we did not find a single font that covered all alphabets in use for this Shared Task, so we had to take some drawbacks and accept rendering of "unknown symbol" placeholders for some languages.

The symbol rendering is handled through the \textit{pygame}\footnote{see \url{https://www.pygame.org/docs/ref/font.html}} library.  More sophisticated alternatives perform anti-aliasing that
nullifies the desired effect of pixel-based comparison. An anti-aliased letter \texttt{a} looks slightly different than the same letter \texttt{ä} with German \textit{Umlaut} added on top, and the resulting patch would contain this noise and therefore be different from the one between e.g.\ \texttt{o} and \texttt{ö}.

\subsection{Equivalence Classes}
After rendering, all resulting patch matrices are grouped by pixel similarity, resulting in a finite number of equivalence classes that can later be used as actions for the transducer. These actions are symmetrical, so that irrespective of lemma and inflection order
we define $p(p(c)) = c$.

Once the patches are grouped, the original pixel representation is discarded so that our data can be arranged as a simple lookup table where patches are represented by numerical indices -- as can be seen in \autoref{fig:patch-lookup}.

We deal with unseen characters during prediction by populating the lookup table over a big portion of the entire Unicode plane, and then filtering the result based on a given input alphabet: We keep all rows of any patch $p$ in the pre-populated table if at least one example of $p$ was observed in the input alphabet.
Although this computation is quite costly, we can still keep runtime demands at a minimum because the whole overview only has to be computed once. Individual languages can then be filtered out "on demand" while holding a complete copy of the Unicode-based lookup table in memory.

\begin{table}
\centering
\begin{tabular}{ccc}
\toprule
Symbol & Patch & Result\\
\midrule
e & 3 & è\\
a & 3 & à\\
& $\cdots$ &\\
o & 17 & ø\\
\bottomrule
\end{tabular}
\caption{Symbol patch lookup table}
\label{fig:patch-lookup}
\end{table}

\section{Enhancing Training Data}
To improve our training on low data quantities, our system can enhance training data by generating artificial samples based only on the existing data. By detecting patterns in words with the same features and generating more data with the same patterns, we assumed that this would aid the network in detecting and applying patterns, such as common prefix and suffix changes.

Similar approaches were taken by submissions for previous CoNLL--SIGMORPHON Shared Tasks. The winning submission \citep{kann-schutze:2016:SIGMORPHON} of the 2016 Shared Task employed data enhancement for the low resource setting. The team of the 2017 submission from \citet{bergmanis:augmenting} used two variants of a sequence autoencoder, with one using lemmas and target forms as inputs and the other using randomly generated strings. The additional training data proved to increase the average performance on development sets. \citet{kann-schutze:2017:K17-20} used several augmentation methods, including a rule based system. \citet{silfverberg-EtAl:2017:K17-20} employ a data augmentation system splitting a word in three parts - inflectional prefix, word stem and inflectional suffix - and then generating new words using existing pre- and suffixes. Further works using data augmentation are provided by \citet{zhou-neubig:2017:K17-20} and \citet{nicolai-EtAl:2017:K17-20}.

\subsection{Basic Enhancement Process}
To generate artificial training samples for a data set, our system sorts the input data into groups of inflections that share the same features. Within each group, it aligns and compares each pair of lemma and inflected form with every other pair, only retaining the common characters at the aligned positions. The different characters are replaced semi-randomly using a language model based on n-grams with one gap each. Finally, these gaps are filled with letters from the dataset based on their frequency (an example is discussed in Section~\ref{sec:enhancer-example}), using the same letters for both the artificial lemma and inflected form. If there are still any gaps left,
more characters are selected based on n-grams from the language model.

\begin{table}
\centering
\begin{tabular}{lccc}
\toprule
n-gram & Letter & Frequency & p\\ \midrule
 \texttt{?ad} & r & $433$ & $0.4446$ \\
 & p & $182$ & $0.1869$\\
 & t & $107$ & $0.1099$\\ 
& \ldots & \ldots & \ldots \\ \midrule
\texttt{?ade} & r & $265$ & $0.5311$\\
 & p & $91$ & $0.1824$\\
 & n & $46$ & $0.0922$ \\
 & \ldots & \ldots & \ldots \\ 
 \bottomrule
\end{tabular}
\caption{Excerpt from the language model for \lang{swedish} (low volume)}
\label{tab:example-langmodel}
\end{table}

The system produces a specified number of words per alignment match. While creating the system we found that more than five enhanced words per match is not beneficial to the end result, with one word generated per match being the best option for most languages. We have also tried adding a constraint regarding the minimum number of occurrences of a pattern necessary to produce artificial words, but found no improvement overall by specifying this minimum support during development.

\subsection{Language Model Example}
\label{sec:enhancer-example}

In \autoref{fig:example-enhancer}, after inserting \texttt{iomm}, one more gap (symbolized by \#) is left to fill. To find an appropriate letter, the current word is compared to the language model's n-grams, starting with $n=5$ and reducing $n$ while shifting the beam from left to right until an n-gram with the corresponding gap is found in the language model. In this case, the longest n-gram found is the 4-gram \texttt{?ade} that can also be seen in \autoref{tab:example-langmodel}. Through using each letter's probability (the frequency of the n-gram in the dataset where the letter replaced the \texttt{?}-symbol) the letter to replace the $?$ gets chosen; in this example it is \texttt{p}.

\begin{table}
\centering
\begin{tabular}{cc}
\toprule
\texttt{skap\textbf{ad}} & \texttt{skapp\textbf{ade}}\\  
\texttt{\#fix\textbf{ad}} & \texttt{\#\#fix\textbf{ade}}\\ \midrule
\texttt{\#\#\#\#\textbf{ad}} & \texttt{\#\#\#\#\#\textbf{ade}} \\
$\Downarrow$ & $\Downarrow$\\
\texttt{iomm\textbf{ad}} & \texttt{iomm\#\textbf{ade}}\\
$\Downarrow$ & $\Downarrow$\\
\texttt{iomm\textbf{ad}} & \texttt{iommp\textbf{ade}}\\
\bottomrule
\end{tabular}
\caption{An example for creating artifical data for \textit{skapad} -- \textit{skapade} (ADJ;DEF), "created"}
\label{fig:example-enhancer}
\end{table}

Theoretically, this system improves with bigger data sets as there are potentially more patterns to be discovered. Unfortunately this also means that for low quantities of data, where enhancement would be most beneficial, the quality of the enhanced data is lower than for higher quantities of data, where it is not as needed.

\section{System Architecture}

\begin{figure}
\includegraphics[width=\linewidth]{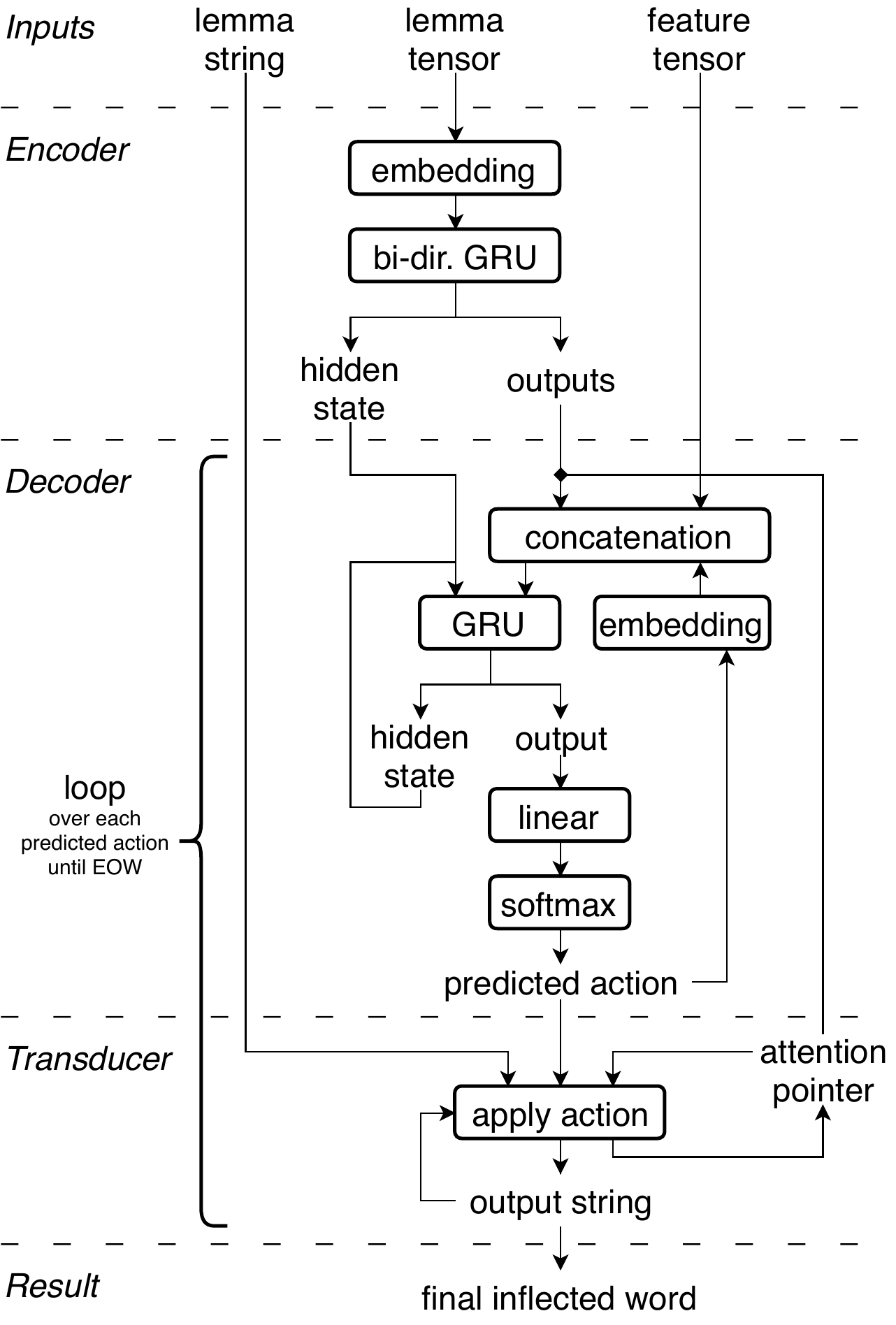}
\caption{System architecture}
\label{fig:arch}
\end{figure}

The system proposed in this work is an encoder-decoder recurrent neural network combined with hard attention and the string-based transducer shown in Section~\ref{sec:transducer}. The architecture is displayed in \autoref{fig:arch}. After processing the inputs through both encoder and decoder the resulting action sequence is applied on the lemma string by the transducer to produce the inflected word.

\subsection{Baseline}
The baseline system that was distributed along with the details for this Shared Task by the organizers is based on pattern matching in strings. It is heavily inspired by the methods proposed in the research of \citet{baseline:Ling2016}.

For any given pair of aligned input lemma and output form, the baseline extracts prefix and suffix rules throughout the entire string, and then greedily applies them on a new input lemma that is to be inflected.
The replacement rules are derived incrementally, so that if multiple rules would match a new sample, the longest one gets applied to produce the most accurate results possible.

Further details about the baseline system can be found in the proceedings of last year's Shared Task \citep{sigmorphon:st2017}, as the architecture is virtually identical.

\subsection{Neural Network Model}
We use the same neural network architecture across all 103 languages and training set sizes (low, medium, high).
The neural network acts as an oracle for the string transducer shown in Section~\ref{sec:transducer}. Its inputs are the lemma of a word and the features of the inflected target form. The outputs correspond to the defined transducer actions (\action{COPY}, \action{PATCH} $p$, \action{MOVE}, \action{EMIT} $s$ and \action{EOW}).

We use an encoder-decoder architecture \citep{encdec:ChoMGBSB14, seq2seq:SutskeverVL14} to transform a sequence of characters into a sequence of transducer actions. The decoder uses hard monotonic attention which has been found beneficial for the task of morphological inflection \citep{hardattention:AharoniG16,hardattention:AharoniEtAl} and allows our system to meaningfully perform \action{COPY} and \action{PATCH} operations.

Both encoder and decoder contain a single gated recurrent unit (GRU) introduced by \citet{encdec:ChoMGBSB14} and character embeddings to obtain a dense numerical representation from each input symbol. The encoder is using a bi-directional GRU whose outputs are summed up from both directions. Since the encoder is uni-directional we only use the forward path of the hidden encoder state to start the decoder. The decoder concatenates the character embedding, attention context and feature tensor as a combined input to the GRU. The decoder GRU output is fed into a linear transform followed by a log softmax layer to obtain the log-likelihoods for each transducer action.

Biases and weights for the GRUs and linear layers are initialized randomly from a uniform distribution $\mathcal{U}(-\sqrt{1/s}, \sqrt{1/s})$ where $s$ is the size of the hidden layer (GRU) or number of input features (linear layer). The embedding weights are initialized randomly from a normal distribution $\mathcal{N}(0, 1)$.

The input lemma is processed at once by the encoder, generating output representations for every input character and hidden state representations for the whole input sequence. 
By using an external loop the decoder produces one transducer action per step. In each step the previous hidden state and output action, inflection features, as well as the attended encoder output is put into the decoder.
Which encoder output is being attended is controlled by the index pointer of the transducer. If the network outputs a \action{MOVE} action, the index pointer is increased so that the decoder will see the next encoder output in the following loop iteration. Actions moving the index pointer beyond the input lemma are discarded.

To improve the prediction performance we implemented a beam-search decoding process. This results in multiple paths out of which the path with the highest probability is selected to produce the final inflected word. An additional transducer state object stores the decoder hidden state, predicted action and its log-likelihood plus the resulting output string for each step and path in the beam.

\subsection{Training}
\label{sec:training}

As the network outputs a sequence of transducer actions, the training targets are not the inflected words but an action sequence which produces the correct inflected form when applied on the lemma. This action sequence is generated by looping over the aligned lemma and inflection word in lockstep. For each character combination the corresponding actions are appended to the new output sequence. The detailed algorithm is described in Section~\ref{sec:oracle-alg}.

Training updates are performed via backpropagation with the Adam optimizer \citep{adam:KingmaB14} using the following parameters: Learning rate $\alpha=0.005$, momentum decays $\beta_1=0.9$, $\beta_2=0.999$, numerical stabilizer $\epsilon=10^{-8}$ and a weight decay (L2 penalty) of $0.001$.

The beam-decoding allows a global normalization of the model according to \citet{globalnorm:AndorAWSPGPC16}. Unfortunately, training the model with global normalization in beam-search failed to converge.
\citet{globalnorm:AndorAWSPGPC16} used pre-training with local normalization to overcome this difficulty, but since we could not find a robust way to switch local to global normalization during training for all 103 languages, we used local normalization only.
Once the correct path falls out of the beam, the log-likelihoods of the correct path build the basis of our custom loss function.

The loss function shown in eq.~\eqref{eq:loss} is based on the locally normalized path probability presented in eq.~(4) of \citet{globalnorm:AndorAWSPGPC16}. It calculates the negated sum over the log-likelihood $l$ of the correct action in each step of the path. Dividing by the natural logarithm of the sequence length $s$ results in a consistent loss magnitude, thus helping the training process to converge more easily. We assume this is the case because we sum up the error across all steps, also punishing the correct predictions if the system was not 100\% confident. The resulting loss $L$ is used to perform the training update back through the entire network.

\begin{equation}
\label{eq:loss}
L = - \frac{\sum_i^s l_i}{\ln{(1+s)}}
\end{equation}

Although local normalization restored convergence of learning, we could not find a significant advantage in using multiple beams during training.
One explanation why our model did not benefit from beam-search might be that it requires many training updates. Punishing the correct steps in the decoding process leads to many updates while with beam-search updates may be too infrequent.

Our final training and evaluation is done with a beam-size of 1. However, the architecture is prepared to utilize both beam-search and global normalization in the future. Training with a single beam and evaluating with multiple beams to find better predictions is also supported. Due to the complex implementation of beam search and combined batching the system works on single training samples by using a batch size of 1.

\subsection{Comparison to previous architectures}
Although our approach follows the "Align and Copy" idea of \citet{cluzh:MakarovRC17} the architectures differ. \citeauthor{cluzh:MakarovRC17} proposed two different models: Hard attention model with copy mechanism (HACM) and hard attention model over edit actions (HAEM). Both contain an encoder-decoder with LSTMs. HACM uses a mixture of character generation and copying probability distribution to implement the copy mechanism.

Our architecture is more similar to HAEM. The latter uses additional LSTMs storing representations of the predicted inflected form, action history and deleted lemma characters. The decoder feeds a concatenation of the feature vector, currently attended encoder output and extra representations through a rectified linear unit followed by a softmax to produce outputs like \action{COPY}, \action{WRITE} and \action{DELETE}. 

\section{Tuning and Evaluation}
\label{sec:tuning_evaluation}
While we used the same architecture for all languages and training set sizes, we performed individual hyperparameter optimization for each language-size-pair. The parameters tested are the hidden size of encoder/decoder ($32, 64, 128$), size of the character embeddings ($8, 16$), whether to use patches or not and what amount of additional training data to hallucinate with the enhancer ($1\times, 5\times$).

During the development we noticed that the results are strongly influenced by the random initialization of the network weights. We therefore tested every parameter combination with five different random seeds to mitigate this issue.
Our final evaluation on the test set used the best parameters we found during the hyperparameter search on the development set for each language-size-pair.

Furthermore, we observed our model sometimes fails to output \action{EOW} and instead either tries to copy non-existent lemma characters or endlessly \action{EMIT}s the same character. The string transducer includes fixes for these issues when the pointer has moved beyond the input lemma. In this case \action{COPY} and \action{PATCH} do not modify the output sequence at all and \action{EMIT} actions cannot append the previously written character again. However, this results in a few missing characters at the end of inflected forms in some corner cases.

\section{Results and Discussion}
\label{sec:results}

Compared to the other \prostst submissions, our system proved to be in the mid-range (top 59\%-67\%). By average accuracy, it improved the most over other submissions for the medium volume datasets.
While the average accuracy increased from $40.3\%$ on the low set by $33.7$ points to $74.0\%$ on the medium set, it improved by only $3.5$ more points from the medium set to $77.5\%$ on the high set. 

An overview over the results on the medium data set is shown in \autoref{tab:medium_averages}. It shows that this system is working exceptionally well on some languages compared to the baseline, such as \lang{Swahili} or \lang{Murrinhpatha}. Likewise, this system performs remarkably worse on some languages, such as \lang{Haida} and \lang{Neapolitan}.

\begin{table}
\centering
\small
\begin{tabular}{llrr}
  \toprule
                           & \textbf{Language}  &\textbf{Ours} & \textbf{BL} \\
\midrule\textbf{Top languages}   & Uzbek              & $100.0$      & $96.0$ \\
                           & Mapudungun         & $100.0$      & $82.0$ \\
                           & Classical-Syriac   & $97.0$       & $99.0$ \\
\textbf{Worst languages}         & Old-Irish          & $6.0$        & $16.0$ \\
                           & Haida              & $16.1$       & $61.0$ \\
                           & Latin              & $21.4$       & $37.6$ \\
\midrule
\textbf{max(Ours - BL)}    & Swahili            & $95$         & $0.0$ \\
                           & Murrinhpatha       & $88.0$       & $0.0$ \\
                           & Zulu               & $81.8$       & $0.1$ \\
\textbf{max(BL - Ours)}    & Neapolitan         & $49.0$       & $94.0$ \\
                           & Haida              & $16.0$       & $61.0$ \\
                           & Latin              & $21.4$       & $37.6$ \\
\bottomrule
\end{tabular}

\vspace{1ex}
\textbf{Above baseline:} $73$  \quad \textbf{avg. diff.:} $20.2$\\
\textbf{Below baseline:} $29$ \quad \textbf{avg. diff.:} $-7.7$

\caption{Results for our system compared to the baseline.
  Languages with the best and worst accuracies and languages that were the furthest above and below the baseline, trained on the medium set and evaluated on the test set.
}
\label{tab:medium_averages}
\end{table}

\subsection{Patches}
Our system is generally able to deduce a meaningful set of patches (that is, a lookup table with more than one trivial entry) for about one third of all languages. While the precise numbers differ per training volume, the overall performance is justified given the font choice discussed in Section~\ref{sec:font-choice}. We could possibly achieve a higher coverage by combining different fonts for different languages, but for us the manual tuning process did not outweigh the work efforts this selection would have required.

We can still observe that out of $42$ languages with patches, our hyperparameter tuning algorithm opted to use patches in $17$ cases on the low environment. While $\frac{17}{42} = 40,4\%$ clearly signifies little to no global improvement, the same fraction rises to $\frac{29}{42} = 69\%$ when evaluating on the medium environment.

In other words, the usefulness of patches rises (among languages that use patches at all) when training our system on larger quantities of data. However at the same time, the selection of which languages actually use patches to achieve maximum accuracy partially differs. Only slightly more than half of the $17$ positively patching languages in the low environment also apply patches on medium, so it is imperative to consider the actual linguistical structures behind the data in order to maximise the benefit of this method.

Lastly, one could combine the NFD system explored in Section~\ref{sec:nfd} with the already implemented font rendering to achieve hybrid patch generation in an effort to maximise its effectiveness. This idea was not pursued further by us and is left open as future work.

\subsection{Data Enhancer}
On low volume, the accuracy on the development set increased for 42 of the enhanced data sets (best of enhancement by 1 / by 5) when compared to the accuracy on the regular data set. For 53 sets they decreased, no matter the enhancement proportion. The probability of these being random observances is $0.3049$ \citep{biostat:zar}.
However, by testing the accuracy of the enhanced and the regular training data for each language on the development set, we can select which languages will be enhanced and which will not. This is part of the hyperparameter search from Section~\ref{sec:tuning_evaluation}. On the low development set, the enhancer is leading to a total improvement of $1.245\%$ in accuracy and a negligible $0.044$ characters in Levenshtein distance, with improvements for single languages of up to $10.8\%$ (\lang{french}). The average improvement is $3.6667\%$.

\subsection{Network Hiccups}
Our system's accuracy is poor on \lang{Haida} and \lang{Neapolitan} compared to other submissions and the baseline. The reason is that for both languages the post-processing used to combat a missing \action{EOW} is often triggered erroneously. The example below shows our system missing the last character in the output because the transducer discards the second identical action to \action{EMIT} an \texttt{a} in this case.
\begin{compactitem}
	\item \texttt{ñíiyä} $\to$ \texttt{ñíiyä'wa\ } (prediction)
    \item \texttt{ñíiyä} $\to$ \texttt{ñíiyä'wa\textbf{a}} (target)
\end{compactitem}
As the inflected words are almost correct, the Levenshtein distance is much lower than the accuracy might indicate. For \lang{Haida} the Levenshtein distance is even significantly lower than the baseline results. In hindsight, it would have been better to replace non-ending predictions with the lemma instead of trying to clean the output as the negative side-effects most likely outweigh any benefits. In the future, a better approach would be to improve the training process by using a dynamic oracle for the target sequence and correctly implementing global normalization with beam-search decoding. These changes are likely to eliminate the need for any post-processing.

Another weakness of our system is the inability to transform a prefix into a suffix or vice versa as shown in the following \lang{German} language example:
\begin{compactitem}
	\item \texttt{\underline{ab}stellen} $\to$ \texttt{stellt \underline{\ \ }} (prediction)
    \item \texttt{\underline{ab}stellen} $\to$ \texttt{stellt \underline{ab}} (target)
\end{compactitem}
This behavior is expected as our neural network works with hard monotonic attention. It would need to store the information within the hidden-state over the whole sequence as it cannot attend the encoder outputs from the beginning again. A cure for this symptom would be to use a model with soft attention -- which in turn cannot meaningfully use \action{COPY} or \action{PATCH} operations on the input lemma.

\subsection{Beam-Decoding}
\begin{figure}
\centering
\includegraphics{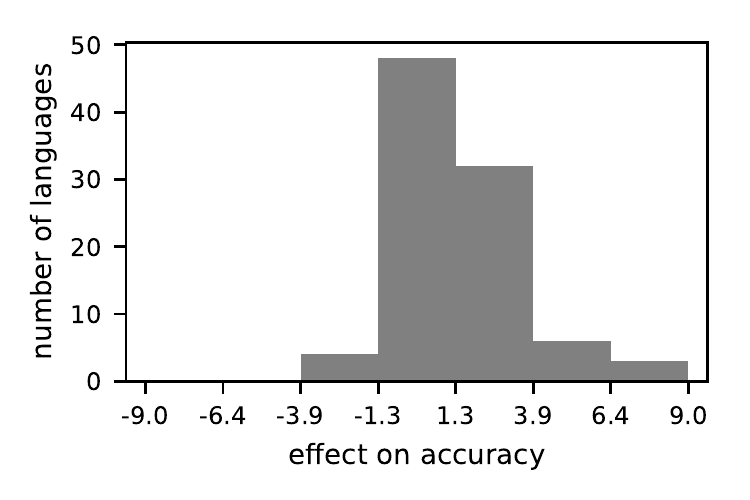}
\caption{Histogram showing the effect of beam size 16 compared to size 1 on the test set (trained on low)}
\label{fig:histBeamLowAcc}
\end{figure}

While we did not use beam-decoding for the official results, we experimented with the evaluation performance after the submission.
\autoref{fig:histBeamLowAcc} shows the number of languages for which beam-decoding with 16 beams makes a difference in comparison to greedy decoding. For half of the languages there are either no or only negligible differences in accuracy. About one third shows a small positive effect. Some languages show a larger accuracy increase while only few languages show a small accuracy decrease.
A binomial test shows that the probability of the increase being random is as low as $2.4 \times 10^{-10}$. Beam-decoding therefore clearly leads to an increase in accuracy which matches the intuition of beam-decoding producing better or equal results compared to greedy decoding.

\section*{Acknowledgements}
We would like to thank the two anonymous reviewers for their help, including making us aware of the Unicode NFD standard.
We also appreciate the work of the organisers to realise the exciting Shared Task.

\bibliography{bib}
\bibliographystyle{acl_natbib}

\appendix

\end{document}